\documentclass[
]{ceurart}

\usepackage[utf8]{inputenc}
\usepackage{graphicx}
\usepackage{amsmath, amssymb}
\usepackage{hyperref}
\usepackage{listings}
\usepackage[numbers]{natbib}
\usepackage{placeins}
\usepackage{caption}

\begin{document}

\copyrightyear{2025}
\copyrightclause{Copyright for this paper by its authors.
Use permitted under Creative Commons License Attribution 4.0
International (CC BY 4.0).}

% \conference{Defactify 4.0: Multimodal Fact-Checking and AI-Generated Content Detection, March 2025, Philadelphia, Pennsylvania, USA}

% \conference{Defactify 4.0: Multimodal Fact-Checking and AI-Generated Content Detection, AAAI Workshop 2025, Philadelphia, Pennsylvania, USA}

\conference{{D}e{F}actify 4.0: Fourth workshop on Multimodal Fact-Checking and Hate Speech Detection, March 2025, Philadelphia, Pennsylvania, USA}

\title{A Comprehensive Dataset for Human vs. AI Generated Image Detection}

\tnotemark[1]
\tnotetext[1]{This document is based on the CEUR-WS template and incorporates topics inspired by the Defactify workshop series.}

\author[1]{Rajarshi Roy}[email=royrajarshi0123@gmail.com]
\author[2]{Ashhar Aziz}
\author[3]{Shashwat Bajpai}
\author[4]{Nasrin Imanpour}
\author[5]{Gurpreet Singh}
\author[6]{Shwetangshu Biswas}
\author[7]{Kapil Wanaskar}
\author[8]{Parth Patwa}
\author[9]{Subhankar Ghosh}
\author[10]{Shreyas Dixit}
\author[1]{Nilesh Ranjan Pal}
\author[4]{Vipula Rawte}
\author[4]{Ritvik Garimella}
\author[14]{Amitava Das}
\author[4]{Amit Sheth}
\author[11]{Gaytri Jena}
\author[12]{Vasu Sharma}
\author[13]{Aishwarya Naresh Reganti}
\author[12]{Vinija Jain}
\author[13]{Aman Chadha}

\address{$^1$Kalyani Government Engineering College, India. $^2$IIIT Delhi, India. $^3$BITS Pilani Hyderabad Campus, India. $^4$University of South Carolina, USA. $^5$IIIT Guwahati, India. $^6$NIT Silchar, India. $^7$San Jos\'{e} State University, USA. $^8$UCLA, USA. $^9$Washington State University, USA. $^{10}$Vishwakarma Institute of Information Technology, India. $^{11}$Gandhi Institute for Technological Advancement, India. $^{12}$Meta AI, USA. $^{13}$Amazon AI, USA. $^{14}$BITS Pilani Goa, India.}

\begin{abstract}
Multimodal generative AI systems like Stable Diffusion, DALL-E, and MidJourney have fundamentally changed how synthetic images are created. These tools drive innovation but also enable the spread of misleading content, false information, and manipulated media. As generated images become harder to distinguish from photographs, detecting them has become an urgent priority. To combat this challenge, we release MS COCOAI, a novel dataset for AI generated image detection consisting of 96000 real and synthetic datapoints, built using the MS COCO dataset. To generate synthetic images, we use five generators: Stable Diffusion 3, Stable Diffusion 2.1, SDXL, DALL-E 3, and MidJourney v6. Based on the dataset, we propose two tasks: (1) classifying images as real or generated, and (2) identifying which model produced a given synthetic image. The dataset is available at \url{https://huggingface.co/datasets/Rajarshi-Roy-research/Defactify_Image_Dataset}.
\end{abstract}

\begin{keywords}
AI-Generated Images \sep Detection Techniques \sep Synthetic Media \sep Generative AI \sep Multimodal AI
\end{keywords}

\maketitle
\vspace{-2.5em}
\begin{center}
    \includegraphics[width=0.8\textwidth]{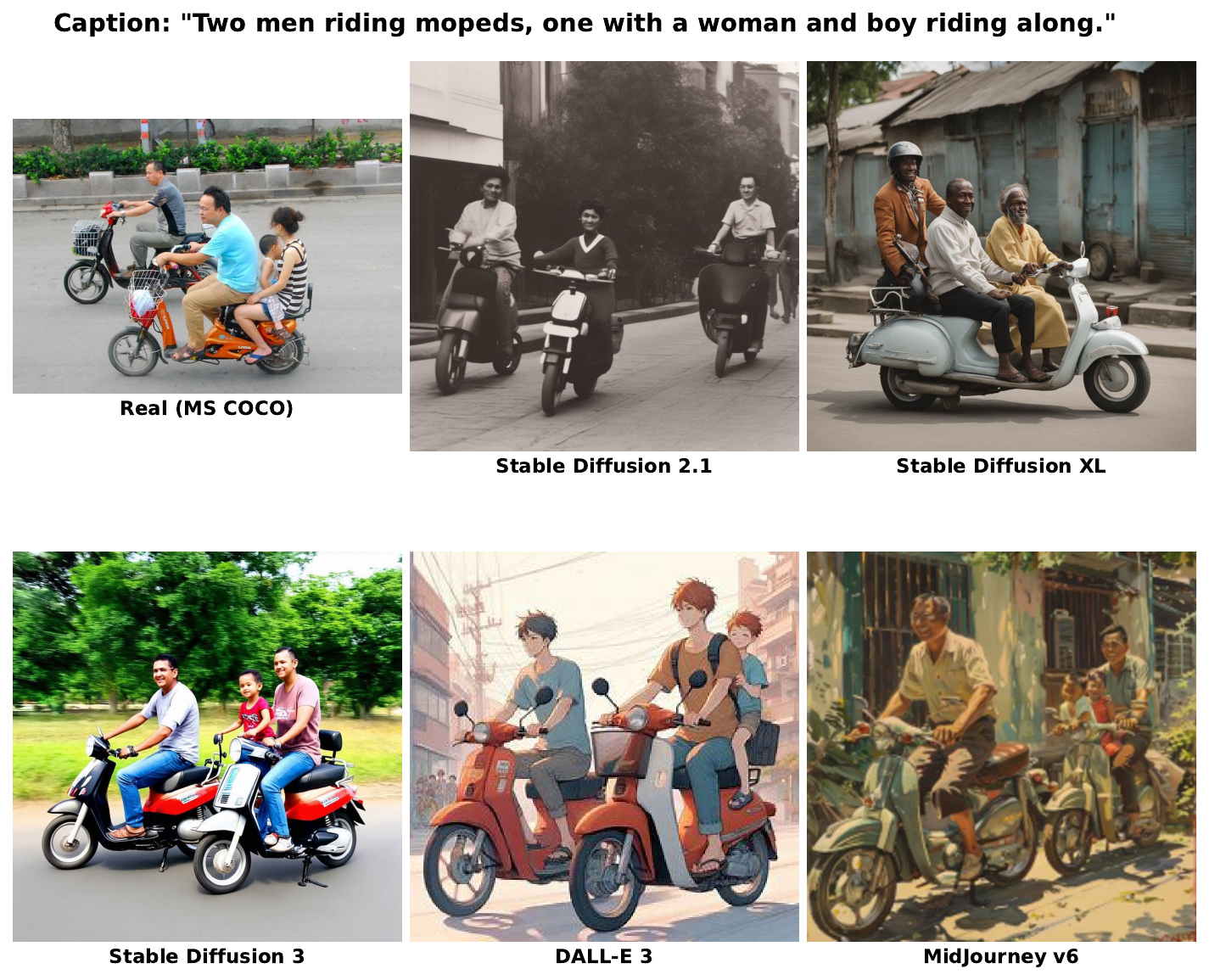}
    \captionof{figure}{Images generated from the same caption. Each model produces visually distinct outputs, highlighting the challenge of AI-generated image detection.}
    \label{fig:teaser}
\end{center}

\section{Introduction}

Generative AI technologies such as Stable Diffusion \cite{rombach2022high}, DALL-E \cite{ramesh2021zero}, and MidJourney \cite{holzmidjourney} have transformed the production of synthetic visual content. These tools, powered by advanced neural architectures, enable diverse applications in fields ranging from advertising and entertainment to design, with prompt quality playing a crucial role in generation outcomes \cite{yang2023prompt}. However, the same innovations that facilitate creative expression also present significant risks when misused. For example, the propagation of misleading or harmful content can disrupt public discourse and undermine trust \cite{mirsky2021creation}.

Recent high-profile incidents have demonstrated the societal impact of AI-generated images, from fabricated depictions that trigger public panic to politically charged visuals intended to sway opinion \cite{akabuogu2025ethics}. The rapid advancement of image generation models has further blurred the line between synthetic and authentic imagery, challenging traditional detection methods and complicating efforts to combat misinformation.

In light of these challenges, there is an urgent need for robust datasets that support the development and evaluation of effective detection techniques. In this paper, we introduce a dataset specifically curated for the detection and analysis of AI-generated images. Our dataset aggregates a diverse collection of images produced by multiple generative models alongside authentic real-world images, and it is enriched with detailed annotations—including the source model, creation timestamp, and relevant contextual metadata. Figure \ref{fig:teaser} provides a sample of our dataset. 

By providing a large-scale, representative benchmark, our dataset aims to advance research in synthetic media detection and foster the development of scalable countermeasures against AI-enabled disinformation. Building upon the foundations laid by initiatives such as the Defactify workshop series \cite{defactify}, this work bridges the gap between academic inquiry and practical implementation, offering a valuable resource for researchers, policymakers, and industry stakeholders committed to safeguarding the integrity of digital information ecosystems. This work complements parallel efforts addressing AI-generated text detection \cite{roy-2025-defactify-dataset-text}.
\section{Related Work}

The rapid growth of generative models has led to highly realistic AI-generated images, making it harder to tell them apart from real images. This section reviews existing datasets and detection methods.

\subsection{AI-Generated Image Datasets}

Several datasets have been introduced for AI-generated image detection:

\begin{itemize}
    \item \textbf{WildFake}: \citet{Hong2024WildFake}  collected fake images from various open-source platforms, covering diverse categories from GANs and diffusion models. However, the uncontrolled collection leads to mixed image quality and no alignment between real and synthetic samples, making it hard to separate generator artifacts from content differences.

    \item \textbf{GenImage}: \citet{Zhu2023GenImage} built a million-scale benchmark with AI-generated and real image pairs. While large in scale, the mainly features older generators and lacks fine-grained model labels, limiting its use for studying modern diffusion models.

    \item \textbf{TWIGMA}: \citet{Chen2023TWIGMA} gathered over 800,000 AI-generated images from Twitter with metadata like tweet text and engagement metrics. While useful for studying real-world sharing patterns, images from social media have compression artifacts and lack controlled generation settings.

    \item \textbf{Fake2M}: \citet{Lu2023Seeing} assembled over two million images and found that humans misclassify 38.7\% of AI-generated images. However, the dataset lacks caption-aligned real-synthetic pairs, preventing controlled studies of how different generators interpret the same text prompt.
\end{itemize}

A shared limitation is the lack of \textit{semantic alignment}, where real and synthetic images share the same text description. This alignment is needed to separate content bias from generation artifacts. Additionally, few datasets cover both open-source (Stable Diffusion) and closed-source (DALL-E, MidJourney) generators, or include perturbations for robustness testing.

\subsection{Detection Methods}

Several approaches have been proposed for detecting AI-generated images:

\begin{itemize}
    \item \textbf{CLIP-Based Detection} \cite{Moskowitz2024Detecting}: Fine-tuning CLIP on mixed real/synthetic data can detect AI-generated images effectively. However, these methods often overfit to specific generators and perform poorly on images from unseen models.

    \item \textbf{Hybrid Feature Methods}: Combining high-level semantic features with low-level noise patterns improves cross-generator performance. Yet, \citet{Yan2024SanityCheck} show that many detectors rely on shortcuts---like scene type or object frequency---rather than true generation artifacts.

    \item \textbf{Frequency-Domain Analysis}: \citet{Corvi_2023_CVPR} show that synthetic images have distinct frequency patterns. While effective for GAN-generated content, diffusion models produce weaker frequency artifacts, requiring new detection approaches.

    \item \textbf{Watermark-Based Detection}: \citet{Guo2024ImageDetectBench} found that watermark-based methods outperform passive detectors under perturbations. However, watermarking needs generator cooperation and fails for models without embedded watermarks.
\end{itemize}

Recent work has also focused on systematic benchmarking of text-to-image generators. \citet{wanaskar2025multimodal} present a unified evaluation framework using metrics such as CLIP similarity, LPIPS, and FID, demonstrating how structured prompts affect generation quality across different architectures.

Beyond model-specific artifact detection, benchmark-oriented evaluations such as the Visual Counter Turing Test (VCT2) \cite{imanpour2025visual} have been proposed to systematically assess detection robustness across generative models, introducing difficulty-aware metrics like the Visual AI Index (VAI).

Our dataset provides caption-aligned real and synthetic images from five modern generators (Stable Diffusion 3 \cite{esser2024scalingrectifiedflowtransformers}, Stable Diffusion 2.1 \cite{rombach2022high}, SDXL \cite{podell2023sdxlimprovinglatentdiffusion}, DALL-E 3 \cite{BetkerImprovingIG}, MidJourney v6 \cite{holzmidjourney}), with model attribution labels and systematic perturbations for robustness evaluation.

\section{Dataset}  
% \section{Dataset Construction}
\label{sec:dataset}

In this section, we describe the dataset creation process and analysis. 

\subsection{Image generation and Annotation}

Our dataset is built upon the MS COCO dataset \cite{lin2014microsoft}, which provides high–quality real images paired with human-written captions. We randomly sample 16k image-caption pairs from the MS COCO dataset. Each caption is used as a textual seed for generating synthetic images using five image-generation models - Stable Diffusion 3  (SD3 ) \cite{esser2024scalingrectifiedflowtransformers}, Stable Diffusion 2.1 (SD 2.1) \cite{rombach2022high}, SDXL \cite{podell2023sdxlimprovinglatentdiffusion}, DALL-E 3 \cite{BetkerImprovingIG}, MidJourney v6 \cite{holzmidjourney}. For every caption, each model produced one synthetic image, resulting in a multi-source collection of AI-generated visual samples.

All real images originate directly from MS COCO, and all captions were written by human annotators as provided by the original dataset. No automated captioning or additional annotation steps are used. All generated images are purely AI-created, whereas the real subset contains only human-captured photographs.

\subsection{Perturbations}
To enable robustness and invariance studies, we create perturbed variants of each generated image using four independent transformations:
\begin{enumerate}
    \item \textbf{Horizontal Flip} – Standard horizontal mirroring of the image.
    \item \textbf{Brightness Reduction} – Image brightness scaled by a factor of $0.5$.
    \item \textbf{Gaussian Noise} – Additive Gaussian noise with standard deviation $\sigma = 0.05$.
    \item \textbf{JPEG Compression} – Image re-encoded using JPEG compression with a quality factor of $50$.
\end{enumerate}
Each perturbation is applied separately, generating distinct augmented versions of each base image. No combined or sequential perturbations are applied.

\subsection{Data Structure}
After collection, the dataset is organized into a standardized schema consisting of the following fields:
\begin{itemize}
    % \item \texttt{id} --- Unique identifier for each sample.
    \item \texttt{image} --- The real or model-generated image.
    \item \texttt{caption} --- The original MS~COCO caption.
    \item \texttt{label\_A} --- Binary label indicating whether the image is real (0) or AI-generated (1).
    \item \texttt{label\_B} --- Categorical label indicating whether an image is real (0) or generated by a specific generative model: Stable Diffusion 2.1 (1), Stable Diffusion XL (2), Stable Diffusion 3 (3), DALL-E 3 (4), or Midjourney 6 (5).
\end{itemize}

All images are produced or collected at the same native resolution, and no resizing or normalization is performed prior to dataset storage. Some examples from the dataset are provided in Figure \ref{fig:teaser}.

\subsection{Data Analysis}

The dataset contains 96,000 image-caption pairs, split into training (42,000), validation (9,000), and test (45,000) subsets. Table~\ref{tab:dataset_stats} summarizes the distribution across sources.

\begin{table}[h]
\centering
\begin{tabular}{|l|c|}
\hline
\textbf{Source} & \textbf{Count} \\
\hline
Real (MS COCO) & 16,000 \\
SD 2.1 & 16,000 \\
SDXL & 16,000 \\
SD 3 & 16,000 \\
DALL-E 3 & 16,000 \\
MidJourney v6 & 16,000 \\
\hline
\textbf{Total} & \textbf{96,000} \\
\hline
\end{tabular}
\caption{Distribution of images by source in the MS COCOAI dataset.}
\label{tab:dataset_stats}
\end{table}

Since all captions originate from MS COCO, they follow its characteristic style: concise, descriptive sentences. Table \ref{tab:caption_stats} provides caption length statistics.

\begin{table}[h]
\centering
\begin{tabular}{|l|c|}
\hline
\textbf{Statistic} & \textbf{Value} \\
\hline
Min words & 7 \\
Max words & 34 \\
Mean words & 10.37 \\
Median words & 10 \\
\hline
\end{tabular}
\caption{Caption length statistics (word count).}
\label{tab:caption_stats}
\end{table}

\begin{figure}[h]
\centering
\includegraphics[width=0.9\textwidth]{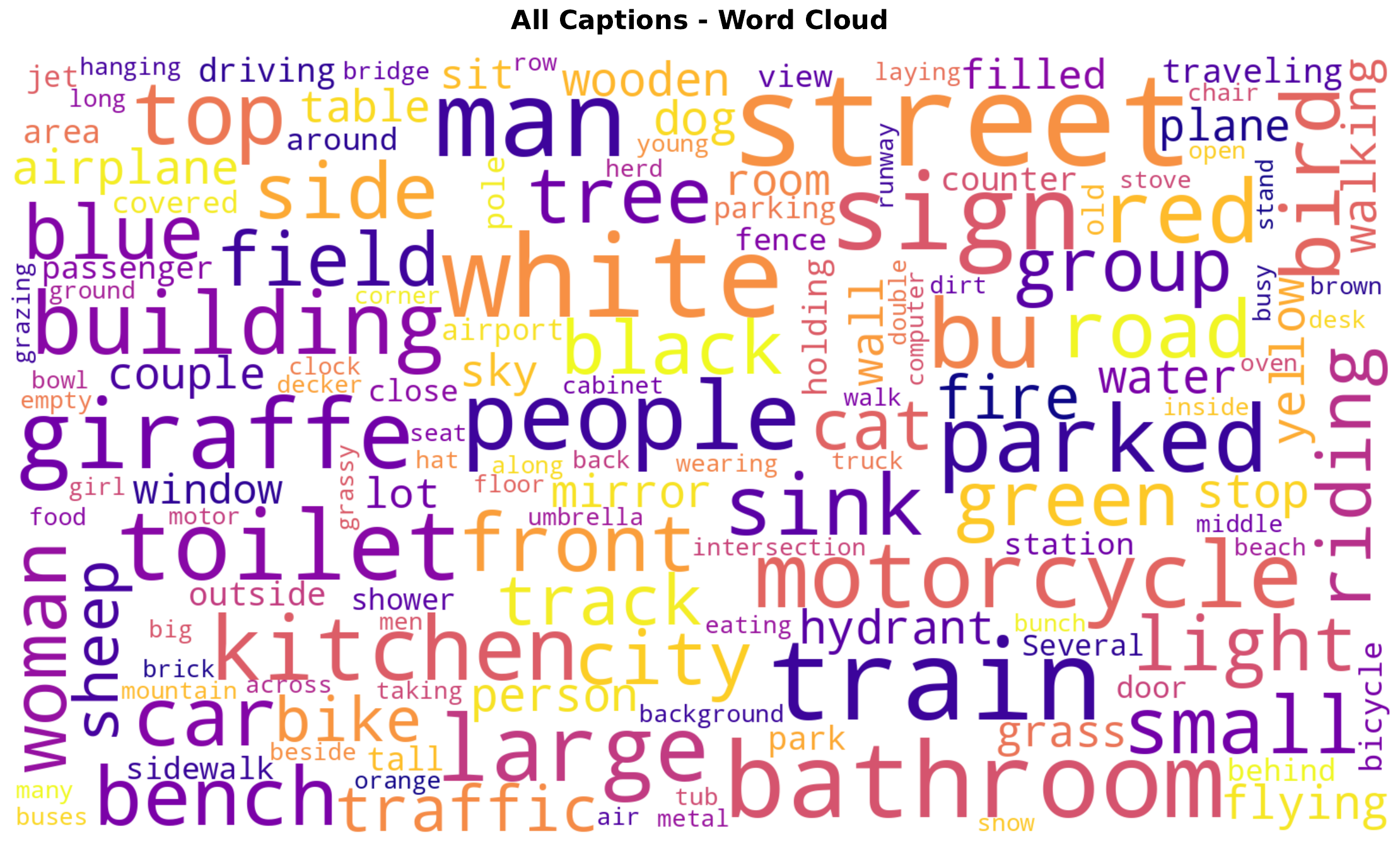}
\caption{Word cloud visualization of all captions in the dataset. Word size corresponds to term frequency, revealing the semantic distribution across the corpus. Dominant terms reflect a comprehensive coverage of everyday scenes, common objects, animals, human activities, and color descriptors.}
\label{fig:wordcloud_all}
\end{figure}

Figure~\ref{fig:wordcloud_all} presents a word cloud of all the captions in our dataset. The visualization reveals the rich semantic diversity, with prominent terms spanning multiple conceptual categories: urban environments (\textit{street}, \textit{building}, \textit{city}, \textit{traffic}), indoor spaces (\textit{kitchen}, \textit{bathroom}, \textit{toilet}, \textit{sink}), wildlife (\textit{giraffe}, \textit{cat}, \textit{dog}, \textit{bird}, \textit{sheep}), transportation (\textit{train}, \textit{bus}, \textit{motorcycle}, \textit{airplane}, \textit{car}), human subjects (\textit{man}, \textit{woman}, \textit{people}, \textit{group}), and descriptive attributes including colors (\textit{white}, \textit{black}, \textit{red}, \textit{blue}, \textit{green}) and spatial relations (\textit{front}, \textit{top}, \textit{side}, \textit{large}, \textit{small}). This broad semantic coverage ensures that generative models are evaluated across diverse visual concepts, reducing the risk of domain-specific biases in detection performance. 

The dataset is available at \url{https://huggingface.co/datasets/Rajarshi-Roy-research/Defactify_Image_Dataset}.

\section{Counter Turing Test (CT2) - AI Generated Image Detection Tasks}  

Based on the dataset, we propose the following 2 tasks: 

\begin{itemize}  
    \item \textbf{Task A (Binary Classification)}: Discern whether a given image is AI-generated or captured in the real world. 
    \item \textbf{Task B (Model Identification)}: Identify the specific generative model (SD 3, SDXL, SD 2.1, DALL-E 3, or MidJourney 6) responsible for producing a given synthetic image. 
\end{itemize}

\section{Baseline}  

To establish a baseline, we train a ResNet-50 classifier using image representations in the frequency domain. These frequency domain representations are generated by applying a preprocessing strategy inspired by the methodology presented in \cite{Corvi_2023_CVPR}. This transformation captures global frequency characteristics that help reveal subtle artifacts often present in synthetic images.

The overall pipeline is illustrated in Figure \ref{fig:baseline_pipeline}. Starting with an input image, we convert it into its frequency domain using a 2D Fourier Transform. The resulting representation is then fed into a ResNet-50 CNN model trained to classify the image into one of the six classes ( real and one class per image generation model).

This baseline allows us to assess the effectiveness of frequency-based features, serving as a point of comparison for more sophisticated techniques.

\begin{center}
    \includegraphics[width=0.9\textwidth]{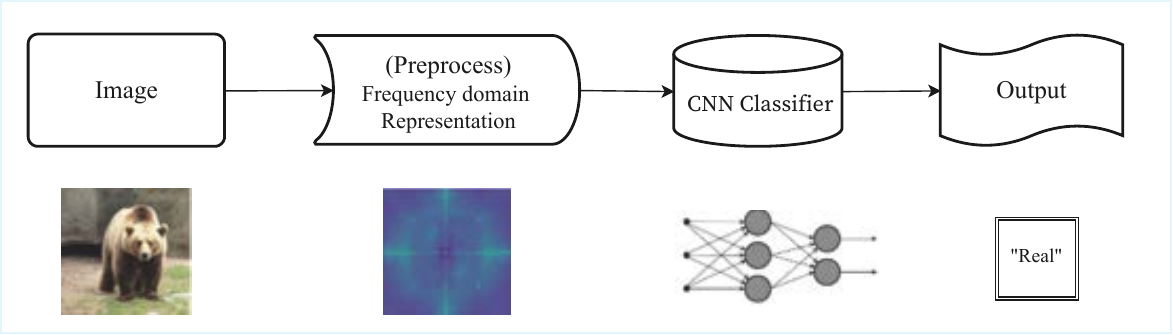} % Adjust width as needed
    \captionof{figure}{\textbf{Baseline workflow.} The input image is first transformed into its frequency domain representation and then passed through a ResNet-50 CNN classifier to predict whether it is real or fake.}
    \label{fig:baseline_pipeline}
\end{center}  

\section{Results}

Baseline performance metrics, given in Table \ref{tab:baseline-results} establish benchmarks for both authenticity detection and model attribution tasks, serving as reference points for subsequent research developments. 

\begin{table}[h]
\centering
\begin{tabular}{|l|c|c|}
\hline
\textbf{Task} & \textbf{Description} & \textbf{Baseline Score} \\
\hline
Task A & Classify each image as either AI-generated or created by a human & 0.80144 \\
\hline
Task B & Given an AI-generated image, determine which specific model produced it & 0.44913 \\
\hline
\end{tabular}
\caption{Baseline results for both tasks.}
\label{tab:baseline-results}
\end{table}

For Task A (binary classification distinguishing AI-generated images from human-created content), the baseline approach achieves a score of 0.80144. For Task B (identifying the specific generative model responsible for producing AI-generated image), the baseline methodology yields a score of 0.44913.

These baseline scores, highlight the substantial difficulty differential between the two tasks, with model attribution proving significantly more challenging than binary authenticity detection. The performance gap demonstrates the increased complexity inherent in multi-class classification scenarios and establishes the dataset as a rigorous benchmark for advancing sophisticated detection and attribution methodologies.
\section{Conclusion}

In this paper, we release a large-scale dataset for AI-generated image detection comprising 96,000 real and synthetic image-caption pairs. A key feature of our dataset is semantic alignment, all synthetic images are generated from the same captions as their real counterparts, enabling controlled studies that separate content bias from generation artifacts.

We propose two tasks based on this dataset: Task A (binary classification of real vs. AI-generated images) and Task B (identifying the specific generative model). Our baseline using ResNet-50 achieves a score of 0.80 on Task A, demonstrating that binary detection is feasible with relatively simple approaches. However, the baseline score of 0.45 on Task B reveals that model attribution remains significantly more challenging.

Future research directions include developing more sophisticated fingerprinting techniques for model attribution, exploring cross-modal learning approaches that leverage caption-image relationships, and improving detector robustness against common image transformations.

\bibliography{references} % Ensure you have a references.bib file
\end{document}